\documentclass[10pt,twocolumn,letterpaper]{article}

\usepackage{wacv}
\usepackage{epsfig}

\usepackage{microtype}
\usepackage{graphicx}
\usepackage{subfigure}
\usepackage{algpseudocode}
\usepackage{booktabs} 
\usepackage{times}
\usepackage{epsfig}
\usepackage{graphicx}
\usepackage{dsfont}
\usepackage{amsmath}
\usepackage{amssymb}
\graphicspath{{./images/}}
\usepackage{tabularx}
\usepackage{makecell}
\usepackage{bm}
\usepackage{multirow} 
\usepackage{rotating}
\usepackage{wrapfig}
\usepackage{subfigure}
\usepackage{amsthm}
\usepackage{thmtools}
\usepackage{thm-restate}
\usepackage{flexisym}
\usepackage{colortbl}
\usepackage{xcolor}
\usepackage{appendix}
\usepackage{mathtools}

\newcommand{\PJ}[2]{\mathbb{P}_{\mathcal{#1}\mathcal{#2}}} 
  
\newcommand{\PI}[2]{\mathbb{P}_{\mathcal{#1}}\otimes\mathbb{P}_{\mathcal{#2}}}

\newcommand{\BV}[1]{\color{black!10!blue}{\em #1}}
\definecolor{ForestGreen}{RGB}{11,102,35}

\newcommand{\rebuttal}[1]{#1}

\usepackage{hyperref}
\hypersetup{colorlinks,linkcolor={red},citecolor={green},urlcolor={blue}}  
\wacvfinalcopy

\def\wacvPaperID{209} 

\ifwacvfinal\fi
\begin{document}

\title{Learning Domain Adaptive Features with Unlabeled Domain Bridges}

\author{Yichen Li\thanks{Both authors contributed equally.}  \\
Boston University\\
{\tt\small liych@bu.edu }
\and
Xingchao Peng\footnotemark[1]\\
Boston University\\
{\tt\small xpeng@bu.edu}
}

\maketitle
\ifwacvfinal\thispagestyle{empty}\fi

\begin{abstract}
Conventional cross-domain image-to-image translation or unsupervised domain adaptation methods assume that the source domain and target domain are closely related. This neglects a practical scenario where the domain discrepancy between the source and target is excessively large. 
In this paper, we propose a novel approach to learn domain adaptive features between the largely-gapped source and target domains with unlabeled domain bridges. 
Firstly, we introduce the framework of Cycle-consistency Flow Generative Adversarial Networks (CFGAN) that utilizes domain bridges to perform image-to-image translation between two distantly distributed domains. Secondly, we propose the Prototypical Adversarial Domain Adaptation (PADA) model which utilizes unlabeled bridge domains to align feature distribution between source and target with a large discrepancy. Extensive quantitative and qualitative experiments are conducted to demonstrate the effectiveness of our proposed models.

\end{abstract}

\section{Introduction}

Supervised machine learning model assumes that the training data and testing data are \textit{i.i.d} sampled from the same distribution, violating a practical learning scenario where the training and testing data are sampled from loosely related domains with heterogeneous distributions, a phenomenon known as \textit{domain shift} or \textit{domain gap} ~\cite{datashift_book2009}. 
Generative models like \textit{CycleGAN}~\cite{CycleGAN2017} tackle the problem of domain shift by generating data across different domains with cycle-consistency loss. \textit{Unsupervised Domain Adaptation} (\textit{UDA})~\cite{adda,DANN,MCD_2018,lsdac} decreases domain shift through aligning the feature distribution between the \textit{source} and \textit{target domains}. However, these state-of-the-art models are designated specifically for adaptation between adjacently-distributed domains. They become inadequate in the more practical scenario where the distributions between the \textit{source} and \textit{target domains} are significantly heterogeneous. 

\begin{figure}[t]
    \centering
    \includegraphics[width=\linewidth]{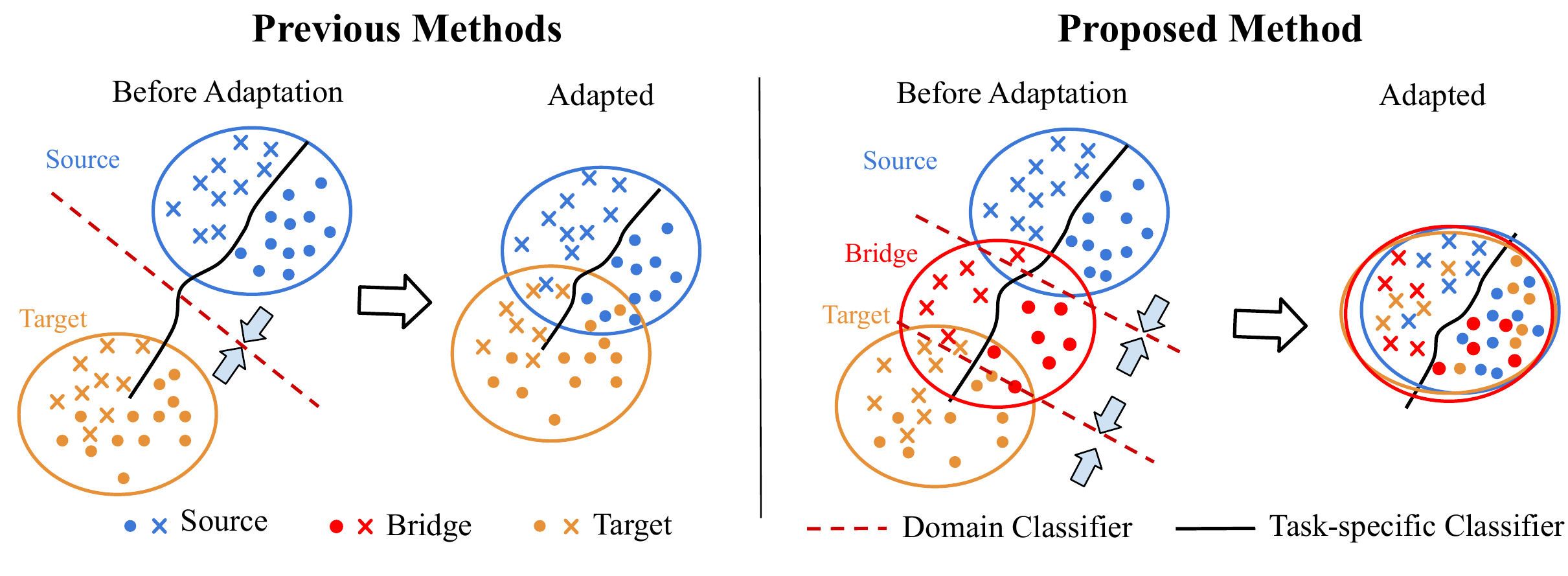}
    \vspace{-0.3cm}
    \caption{We study domain adaptation between two distantly distributed domains. Comparison of previous and the
proposed methods using \textit{domain bridge}. \textbf{Left}: Previous methods directly map \textit{source} to \textit{target}, and only achieves limited alignment in the large-gap domain adaptation scenario.  \textbf{Right}: We propose to utilize an existing intermediate domain to bridge the knowledge transfer from \textit{source} to \textit{target} domain in the significant domain shift scenario. }

    \label{figure_1}
    \vspace{-0.2cm}
\end{figure}

In this paper, we consider a learning scenario where domain shift between source and target domains is significantly large. The main challenges of this learning scenario are: (\textbf{1}) the excessive domain discrepancy hampers the effectiveness of mainstream cross-domain machine learning models, as showed in Figure~\ref{figure_1}; 
(\textbf{2}) the class-irrelevant features between the source and target domains lead to significant \textit{negative transfer}~\cite{pan2010survey}, which 
occurs frequently when the two domains are highly heterogeneous.

The mainstream domain-adaptive image-translation methods based on conditional Generative Adversarial Networks (GANs)~\cite{isola2017image, wang2018high, shrivastava2017learning, bousmalis2017unsupervised, taigman2016unsupervised, benaim2017one} 
assume that there exists a large amount of paired data. However, such data are hard to acquire. \textit{Cycle-consistency} loss~\cite{CycleGAN2017} is then proposed to enable conditional GANs to generate cross-domain images using 
unpaired data. Inspired by \cite{CycleGAN2017}, \textit{multi-domain} image-to-image generation frameworks~\cite{ufdn,choi2018stargan} are proposed to generate images in the presence of multiple domains. However, 
these methods make the 
assumption that the \textit{source} and the \textit{target domains} are closely distributed.
Empirical evidences~\cite{CycleGAN2017,UNIT} show their performances can be severely impeded when domain shift is significant. For example, when trying to adapt a source domain to a distant target domain, the CycleGAN model generates many undesired artifacts ~\cite{CycleGAN2017}. 

Unsupervised domain adaptation models align the source domain to the target by minimizing the Maximum Mean Discrepancy~\cite{long2015,JAN,ddc} through aligning high-order moments~\cite{cmd, peng2017synthetic} or adversarial training~\cite{DANN, adda}. However, these methods are devised specifically for one-to-one domain alignment, while the more practical multi-source domain adaptation~\cite{lsdac,xu2018deep} methods transfer the knowledge learned from multiple labeled source domains to a target domain. These existing methods are designed under the
assumption that the aligned domains
possess similar and adjacent distributions. Their performances are severely 
hampered when the domain shift is remarkable~\cite{lsdac}. {\rebuttal{The \textit{Distant Domain Transfer Learning} (DDTL) framework~\cite{tan2017distant} proposes to align two domains with unrelated concepts by learning intermediate concepts gradually. However, their ``distant domain" is defined based on the \textit{concept shift} ($p_{S}(y|x)\not=p_{T}(y|x)$), which cannot tackle the domain adaptation setting proposed in our paper.}}

We postulate a novel domain adaptation solution 
to learn domain adaptive features between domains with extreme domain shift. 
Our intuition is to leverage an intermediate domain to 
facilitate knowledge transfer from two distant 
domains, as showed in Figure~\ref{figure_1}. 

For \textbf{image-to-image translation}, we propose Cycle-consistency Flow Generative Adversarial Networks (CFGAN) to perform image translation between two largely-gapped domains, \textit{e.g.} from quick draw to real images from the \textit{DomainNet} dataset. Inspired by Cycle-GAN~\cite{CycleGAN2017}, our model uses cycle-consistency loss to translate images between (\textit{source}, \textit{bridge}) domain pairs, and then between (\textit{bridge}, \textit{target}) domain pairs. For \textbf{unsupervised domain adaptation}, we devise Prototypical Adversarial Domain Adaptation framework, which utilizes \textit{bridge} domain to facilitate knowledge transfer from the \textit{source domain} to a distant \textit{target domain}, as showed in Figure~\ref{figure_1}. Specifically, we leverage the Prototypical Matching Network (PMN) to align the (\textit{source}, \textit{bridge}) and (\textit{bridge}, \textit{target}) domain pairs with Maximum Mean Discrepancy~\cite{gretton2007kernel} loss. To enhance domain adaptation, we utilize the feature disentangling component to dispel the class-irrelevant features, aiming to reduce the potential negative transfer.

The main contributions of this paper can be highlighted as follows: \textbf{(1)} we propose a new domain adaptive learning paradigm where the domain shift between the source and target domain is remarkable; \textbf{(2)} we propose a CFGAN framework to perform image-image translation between domains with remarkable domain shift; \textbf{(3)} we propose a novel PADA approach to tackle the unsupervised domain adaptation task with significant domain shift.
\section{Related Work}

\noindent \textbf{Image-to-Image Translation} Image-to-image translation aims to generate images similar to the ones from the target domain by constructing a mapping function between the \textit{source} and \textit{target domains}. Isola \textit{et al}~\cite{isola2017image} proposes the first unified framework for image-to-image translation based on conditional GANs, which has been extended to generate high-resolution images by Wang \textit{et al}~\cite{wang2018high}. Inspired by this approach, some works are proposed focusing on preserving certain properties of the source domain data, such as pixel information~\cite{shrivastava2017learning, bousmalis2017unsupervised}, semantic cues~\cite{taigman2016unsupervised}, pairwise sample distances~\cite{benaim2017one}, or category labels~\cite{bousmalis2017unsupervised}. A significant drawback of these methods is that the limited availability and accessibility of paired data. 
To tackle this problem, CycleGAN~\cite{CycleGAN2017} introduces a cycle-consistency loss to recover the original images using a cycle of translation and reverse translation. 
However, these methods assume that the domain gap between the source and target is relatively small. In contrast, we consider the scenario where the domains to adapt are significantly gapped and propose to utilize an intermediate domain to bridge the source and target domains.

\noindent \textbf{Multi-Domain Learning} Multi-Domain Learning aims to incorporate visual cues from different domains to a single model. Inspired by the early theoretical analysis~\cite{ben2010theory,Mansour_nips2018,crammer2008learning}, multi-domain learning has facilitated the applications in object recognition~\cite{xu2018deep, lsdac}, event recognition~\cite{duan2012exploiting}, and natural language processing~\cite{joshi2013s}. Liu \textit{et al}~\cite{ufdn} and Choi \textit{et al} propose using applied generative model to generate images with the features learned from multiple domains~\cite{choi2018stargan}. {\rebuttal{Gholami \textit{et al}~\cite{gholami2018unsupervised} proposes unsupervised multi-target domain adaptation, assuming that target domain labels are provided while training. In contrast, Peng \textit{et al}~\cite{DAL_DADA} and Chen \textit{et al}~\cite{chen2019blending} introduce a blending-target domain adaptation setting when the domain labels are absent.}} These methods assume that the source and target domains are closely related. In this paper, we propose to bridge knowledge transfer between two distant domains with unlabeled intermediate domains to tackle excessive domain shifts. 

Recently, the idea of bridge domain has gained favor in the field of \textit{multi-domain learning}. Wei \textit{et al}~\cite{wei2018person} proposes to bridge the domain gap for person re-identification with Person Transfer Generative Adversarial Network. More recently, Gong \textit{et al}~\cite{gong2018dlow} proposes to generate multiple intermediate domains using the \textit{source} and \textit{target domains} to decrease domain shift. Our approach differs from these methods in two aspects: (\textbf{1}) our approach is devised specifically to tackle the significantly large domain shift, (\textbf{2}) instead of directly synthesizing bridge domains using the \textit{source} and \textit{target domains}, we leverage an existing third domain to bridge two distant source and target domains. {\rebuttal{Tan \textit{et al}~\cite{tan2017distant} proposes DDTL to bridge the distant domain shift with selective learning algorithm. Our paper differs from DDTL in the following aspects: (1) the ``domain'' in DDTL corresponds to a concept or class ($p(y|x)$) for a specific classification problem, such as face or airplane recognition from images. In our paper, the remarkable domain shift refers to the marginal domain gap (p(x)) between the source and target domain, such as the sketch airplane and photo-realistic airplane. (2) The DDTL framework transfer knowledge learned from rich-labeled domain to coarse-label domain. In our paper, the target and bridge domains are unlabeled.}}

\noindent \textbf{Domain Adaptation} Domain adaptation is a specialized form of transfer learning~\cite{pan2010survey}, which aims to learn a model from a source domain that can generalize to a different but related target domain. The domain adaptation problem can be classified into different categories based on the number of target samples. By denoting the number of target samples as $N^t$ and the number of the labeled target samples as $N^{tl}$, we can categorize domain adaptation into (1) \textit{unsupervised DA}~\cite{long2015,DANN,adda,CycleGAN2017},  \textit{if $N^{tl}$=0}; (2) \textit{supervised DA}~\cite{Tzeng_2015_ICCV, koniusz2017domain}, \textit{if $N^{tl}$=$N^t$}; (3) \textit{semi-supervised DA}~\cite{guo2012cross, gong2012geodesic, yao2015semi}, \textit{otherwise}. Specifically, Long \textit{et al}~\cite{long2015} leverages multi-kernel maximum mean discrepancy (MK-MMD)~\cite{gretton2007kernel} to minimize the domain shift, without the supervision from target domain. Tzeng \textit{et al}~\cite{Tzeng_2015_ICCV} proposes to facilitate domain transfer by a soft label distribution matching loss. Gong \textit{et al}~\cite{gong2012geodesic} proposes Geodesic Flow Kernel to bridge two domains by integrating an infinite number of subspaces that characterize changes in geometric and statistical properties. One limitation of the aforementioned methods is that they take into assumption the domain shift between the source and target domains is relatively small. In contrast, we consider a realistic scenario where the domain gap is large enough that these methods have very limited application.

\section{Domain Discrepancy}
We define a bi-directional domain discrepancy 
between two domains to facilitate our comparison of the distances between domains. Previous works~\cite{long2015,JAN} have applied KL-divergence~\cite{kullback1951information} or Maximum Mean Discrepancy~\cite{gretton2007kernel} as domain distance measure. However, KL-divergence is not symmetrically defined and most previous works apply MMD in kernel reproducing Hilbert space. Instead, we propose to utilize $\mathcal{H}\Delta\mathcal{H}$~\cite{ben2010theory} divergence to evaluate the domain shift.

 \noindent \textbf{Notation} Let $\mathcal{D}_s$\footnote{In this literature, the calligraphic $\mathcal{D}$ denotes data distribution, and italic \textit{D} denotes domain discriminator.} and $\mathcal{D}_t$ denote source and target distribution on input space $\mathcal{X}$ and a ground-truth labeling function $g:\mathcal{X}\to\{0, 1\}$. A \textit{hypothesis} is a function $h:\mathcal{X}\to\{0,1\}$ with the \textit{error} w.r.t the ground-truth labeling function $g$: $\epsilon_S(h, g) := \mathbb{E}_{\mathbf{x}\sim\mathcal{D}_s}[|h(\mathbf{x}) - g(\mathbf{x})|]$. We denote the risk and empirical risk of hypothesis $h$ on $\mathcal{D}_s$ as $\epsilon_S(h)$ and $\widehat{\epsilon}_S(h)$. Similarly, the risk and empirical risk of $h$ on $\mathcal{D}_t$ are denoted as $\epsilon_T(h)$ and $\widehat{\epsilon}_T(h)$. The $\mathcal{H}$-divergence~\cite{ben2010theory} between two distributions $\mathcal{D}$ and $\mathcal{D}'$ is defined as: $d_{\mathcal{H}}(\mathcal{D}, \mathcal{D}'):= 2\sup_{A\in\mathcal{A}_{\mathcal{H}}}|\Pr_{\mathcal{D}}(A) - \Pr_{\mathcal{D}'}(A)|$, where $\mathcal{H}$ is a hypothesis class for input space $\mathcal{X}$, and $\mathcal{A}_{\mathcal{H}}$ denotes the collection of subsets of $\mathcal{X}$ that are the support of some hypothesis in $\mathcal{H}$. The $\mathcal{H}\Delta\mathcal{H}$ divergence of a measurable hypothesis class $\mathcal{H}$ is defined as: $\mathcal{H}\Delta\mathcal{H}:=\{h(\mathbf{x})\oplus h'(\mathbf{x}))|h,h'\in\mathcal{H}\}$, ($\oplus$: the XOR operation). In other words, every hypothesis $g \in \mathcal{H}\Delta\mathcal{H}$ is the set of disagreements between two hypotheses in $\mathcal{H}$. {\rebuttal{We define the distance between two domains in the hypothesis $\mathcal{H}$ as: $\text{\fontfamily{qcr}\selectfont dist} (\mathcal{D}_s, \mathcal{D}_t) = d_{\mathcal{H}\Delta\mathcal{H}}(\mathcal{D}_s, \mathcal{D}_t)$}}

\begin{figure}[t]
    \centering
    \vspace{0.1cm}
    \includegraphics[width=0.8\linewidth]{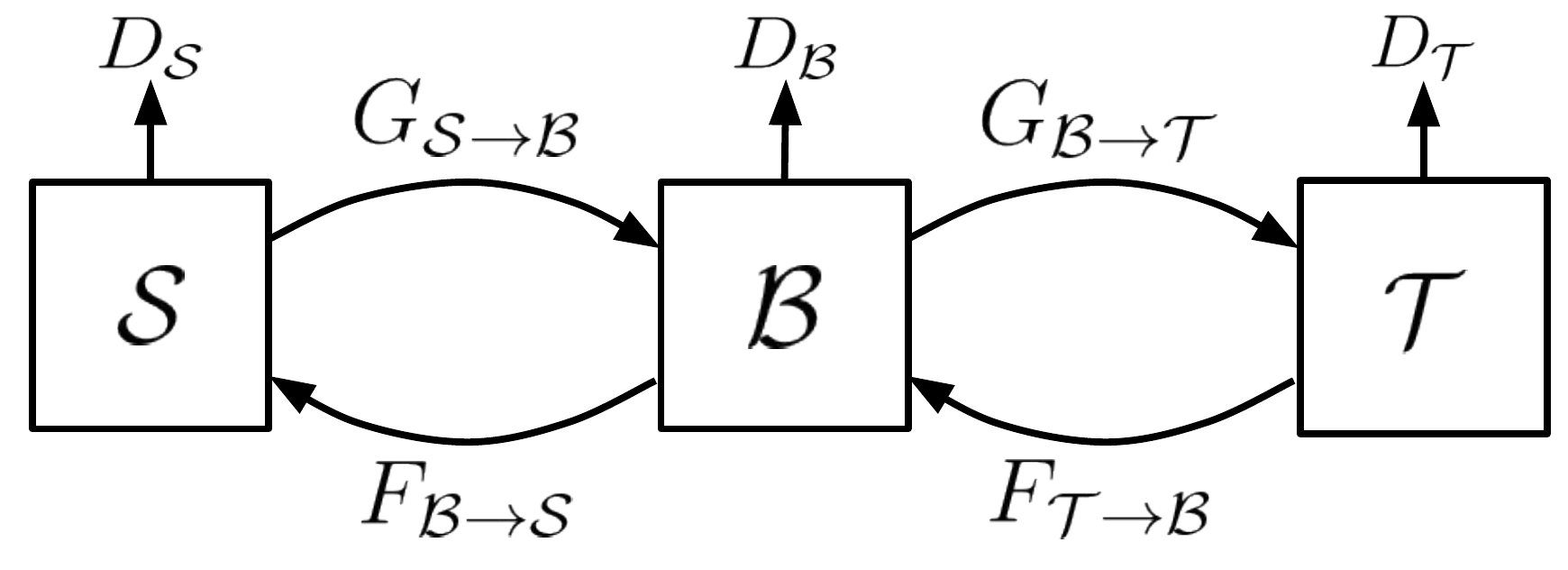}
    \vspace{-0.2cm}
    \caption{We introduce a bridge domain $\mathcal{B}$ to transfer feature knowledge learned on the source domain $\mathcal{S}$ to the target domain $\mathcal{T}$. Our model comprises multiple interconnected Cycle-GAN architectures, including an initial translation from $\mathcal{S}$ to $\mathcal{B}$ and a subsequent mapping from $\mathcal{B}$ to $\mathcal{T}$. Specifically, discriminator $D_{\mathcal{B}}$ encourages generator $G_{\mathcal{S} \rightarrow \mathcal{B}}$ to translate $\mathcal{S}$ into outputs indistinguishable in the bridge domain $\mathcal{B}$. Subsequently, $G_{\mathcal{B} \rightarrow \mathcal{T}}$ translates the results in $\mathcal{B}$ to the desired target domain $\mathcal{T}$. }

    \label{fig_cf_gan_overview}
    \vspace{-0.2cm}
\end{figure}
\begin{figure*}[t]
    \centering
    \vspace{-0.4cm}
    \includegraphics[width=\linewidth]{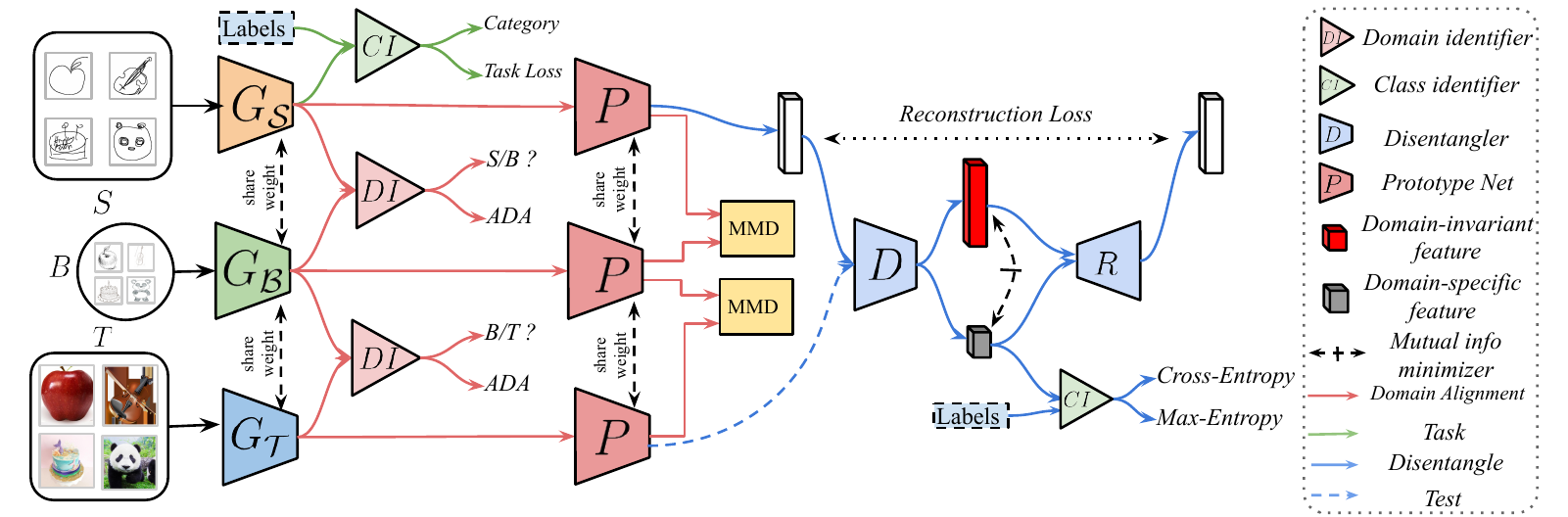}
    \vspace{-0.6cm}
    \caption{{\bf PADA framework.} Our model Prototypical Adversarial Domain Adaptation (PADA) comprises of three components: (\textbf{1}) Adversarial Domain Alignment (ADA), (\textbf{2}) prototypical matching network, (\textbf{3}) disentangle component. In the training phase, the feature extractor $G$ maps the input data to a latent feature space. We then train the domain identifier with the extracted features and domain labels. Consequently, we confuse the domain identifier with adversarial loss. Next, we leverage a prototypical matching network to create the prototypes and align the prototype space with Maximum Mean Discrepancy (MMD) loss~\cite{gretton2007kernel}.In the last step, we disentangle the prototype feature into \textit{class-irrelevant} features $f_{ci}$ and \textit{domain-invariant} $f_{di}$ features with a disentangling auto-encoder. We first utilize a disentangler $\mathbf{D}$ and the \textit{K}-way class identifier $CI$ to correctly predict the labels. We then train the disentangler $\mathbf{D}$ to fool the class identifier by generating \textit{class-irrelevant} features $f_{ci}$. The reconstructor $R$ takes $f_{ci}$ and $f_{di}$ to recover the original features with a reconstruction loss.
    }

    \label{fig_PADA_overview}
    \vspace{-0.4cm}
\end{figure*}

\section{CFGAN: CycleFlow GAN}
We first recapitulate the CycleGAN method~\cite{CycleGAN2017} then describe the details of our CFGAN model. 

\noindent \textbf{CycleGAN Recap}
CycleGAN~\cite{CycleGAN2017} proposes to translate an image from a source domain $\mathcal{S}$ to a target domain $\mathcal{T}$ with two generator-discriminator pairs, $\{G, D_{\mathcal{T}}\}$ and $\{F, D_{\mathcal{S}}\}$. For generator $G$ and its associated discriminator $D_{\mathcal{T}}$, the training loss is
\begin{align}
    \mathcal{L}_{\mathcal{T}adv}(G,D_{\mathcal{T}},\mathcal{S},&\mathcal{T}) = \ \mathbb{E}_{z \sim p_{\text{data}}(\mathcal{T})}[\log D_{\mathcal{T}}(z)] \nonumber \\
   +& \ \mathbb{E}_{x \sim p_{\text{data}}(\mathcal{S})}[\log (1-D_{\mathcal{T}}(G(x))],
\end{align}
where $x$ and $z$ are data sampled from $\mathcal{S}$ and $\mathcal{T}$, respectively. The $G$ tries to generate images $G(x)$ that look similar to images from domain $\mathcal{T}$, while $D_{\mathcal{T}}$ aims to distinguish between translated images $G(x)$ and real samples $z$. The adversarial loss for the mapping function $F:\mathcal{T}\rightarrow\mathcal{S}$ and its discriminator $D_{\mathcal{S}}$ is similarly defined as
\begin{align}
    \mathcal{L}_{\mathcal{S}adv}(F,D_{\mathcal{S}},\mathcal{T},&\mathcal{S}) = \ \mathbb{E}_{x \sim p_{\text{data}}(\mathcal{S})}[\log D_{\mathcal{S}}(x)] \nonumber \\
   +& \ \mathbb{E}_{z \sim p_{\text{data}}(\mathcal{T})}[\log (1-D_{\mathcal{S}}(G(z))].
\end{align}

To recover the original image after a cycle of translation $G:\mathcal{S}\rightarrow\mathcal{T}$ and reverse translation $F:\mathcal{T}\rightarrow\mathcal{S}$, CycleGAN leverages a cycle-consistency loss:
\begin{align}
    \mathcal{L}_{\text{cyc}}(G, F) =  & \ \mathbb{E}_{x\sim p_{\text{data}}(\mathcal{S})}[\|F(G(x))-x\|_1] \nonumber \\ 
    + &\ \mathbb{E}_{z\sim p_{\text{data}}(T)}[{\|G(F(z))-z\|}_1].
\end{align}

\subsection{CycleFlow GAN}
The performance of CycleGAN is hampered when the domain shift is considerably large, as shown in the failure cases of ~\cite{CycleGAN2017}. To tackle this problem, we propose a new framework named CycleFlowGAN (CFGAN). Specifically, we introduce an intermediate domain $\mathcal{B}$ to bridge the source domain $\mathcal{S}$ and the target domain $\mathcal{T}$. The bridge domain is selected under the following constraints: {\fontfamily{qcr}\selectfont dist}($\mathcal{S}$, $\mathcal{B}$) \textless {\fontfamily{qcr}\selectfont dist}($\mathcal{S}$, $\mathcal{T}$) and {\fontfamily{qcr}\selectfont dist}($\mathcal{B}$, $\mathcal{T}$) \textless {\fontfamily{qcr}\selectfont dist}($\mathcal{S}$, $\mathcal{T}$). CFGAN contains two translation generator-discriminator pairs ($G_{\mathcal{S}\rightarrow\mathcal{B}}$, $D_{\mathcal{B}}$) and ($G_{\mathcal{B}\rightarrow\mathcal{T}}$, $D_{\mathcal{T}}$). {\rebuttal{As shown in Figure~\ref{fig_cf_gan_overview}, the whole pipelines can be symmetrically trained in an end-to-end manner. The generator pairs ( $G_{\mathcal{S}\rightarrow\mathcal{B}}$, $G_{\mathcal{B}\rightarrow\mathcal{T}}$) are designed to transfer the source to the target and ($F_{\mathcal{T}\rightarrow\mathcal{B}}$, $F_{\mathcal{B}\rightarrow\mathcal{S}}$) are responsible to transform the target to source domain.}} The training loss is:
\begin{align}
    \mathcal{L}_{\mathcal{T}adv}&(G_{\mathcal{S}\rightarrow\mathcal{B}},D_{\mathcal{B}},G_{\mathcal{B}\rightarrow\mathcal{T}},D_{\mathcal{T}},\mathcal{S},\mathcal{B},\mathcal{T}) = \nonumber \\ 
    &\mathbb{E}_{y \sim p_{\text{data}}(\mathcal{B})}[\log D_{\mathcal{B}}(y)] + \lambda \mathbb{E}_{z \sim p_{\text{data}}(\mathcal{T})}[\log D_{\mathcal{T}}(z)] \nonumber \\
   &+ \mathbb{E}_{x \sim p_{\text{data}}(\mathcal{S})}[\log (1-D_{\mathcal{B}}(G_{\mathcal{S}\rightarrow\mathcal{B}}(x))] \nonumber \\
   & + \lambda \mathbb{E}_{y \sim p_{\text{data}}(\mathcal{B})}[\log (1-D_{\mathcal{T}}(G_{\mathcal{B}\rightarrow\mathcal{T}}(G_{\mathcal{S}\rightarrow\mathcal{B}}(x)))].\nonumber
\end{align}

where $x$, $y$ and $z$ are data sampled from $\mathcal{S}$, $\mathcal{B}$ and $\mathcal{T}$, respectively. $\lambda$ is the trade-off parameter between two generators. {\rebuttal{We also define the similar loss function $\mathcal{L}_{\mathcal{S}adv}(F_{\mathcal{T}\rightarrow\mathcal{B}},D_{\mathcal{B}},F_{\mathcal{B}\rightarrow\mathcal{S}},D_{\mathcal{S}},\mathcal{T},\mathcal{B},\mathcal{S})$ for the reverse translation.}}

To recover the original images after a flow of cycle translation $G_{\mathcal{S}\rightarrow\mathcal{B}}$, $G_{\mathcal{B}\rightarrow\mathcal{T}}$ and reverse translation $F_{\mathcal{T}\rightarrow\mathcal{B}}$, $F_{\mathcal{B}\rightarrow\mathcal{S}}$, we introduce the cycle-consistency loss for CFGAN as:
\begin{align}
    \mathcal{L}_{\text{cyc}}&(G_{\mathcal{S}\rightarrow\mathcal{B}},G_{\mathcal{B}\rightarrow\mathcal{T}},F_{\mathcal{T}\rightarrow\mathcal{B}},F_{\mathcal{B}\rightarrow\mathcal{S}}) =  \nonumber \\
    &\ \mathbb{E}_{y\sim p_{\text{data}}(B)}[{\|G_{\mathcal{S}\rightarrow\mathcal{B}}(F_{\mathcal{B}\rightarrow\mathcal{S}}(y))-y\|}_1]. \nonumber \\
    &+\mathbb{E}_{x\sim p_{\text{data}}(\mathcal{S})}[\|F_{\mathcal{B}\rightarrow\mathcal{S}}(G_{\mathcal{S}\rightarrow\mathcal{B}}(x))-x\|_1] \nonumber \\ 
    &+ \lambda \mathbb{E}_{z\sim p_{\text{data}}(T)}[{\|G_{\mathcal{B}\rightarrow\mathcal{T}}(F_{\mathcal{T}\rightarrow\mathcal{B}}(z))-z\|}_1].     \nonumber  \\
    &+\lambda \mathbb{E}_{x\sim p_{\text{data}}(\mathcal{S})}[\|F_{\mathcal{T}\rightarrow\mathcal{B}}(G_{\mathcal{B}\rightarrow\mathcal{T}}(G_{\mathcal{S}\rightarrow\mathcal{B}}(x)))-G_{\mathcal{S}\rightarrow\mathcal{B}}(x)\|_1]. \nonumber 
\end{align}

\section{Domain Adaptation with Domain Bridge}
We define the task of domain adaptation with domain bridge as follows: Given a \emph{source} domain $\mathcal{D}_s = \{(\mathbf{x}_i^s,y^s_i)\}_{i=1}^{n_s}$ with $n_s$ labeled examples, the goal is to minimize risk on unlabeled target domain $\mathcal{D}_t=\{\mathbf{x}_i^t\}_{i=1}^{n_t}$ with $n_t$ examples. We assume that {\fontfamily{qcr}\selectfont dist}($\mathcal{D}_s$, $\mathcal{D}_t$) is significantly large where the conventional domain adaptation methods become inadequate. To tackle this problem, we introduce an intermediate domain $\mathcal{D}_b = \{\mathbf{x}_i^b\}_{i=1}^{n_b}$ to facilitate knowledge transfer from $\mathcal{D}_s$ to $\mathcal{D}_t$. We carefully select $\mathcal{D}_b$ under the following constraints: {\fontfamily{qcr}\selectfont dist}($\mathcal{D}_s$, $\mathcal{D}_b$) \textless {\fontfamily{qcr}\selectfont dist}($\mathcal{D}_s$, $\mathcal{D}_t$) and {\fontfamily{qcr}\selectfont dist}($\mathcal{D}_b$, $\mathcal{D}_t$) \textless {\fontfamily{qcr}\selectfont dist}($\mathcal{D}_s$, $\mathcal{D}_t$).  Empirically, we want to minimize the target risk ${\epsilon_t}\left( \theta  \right) = {\Pr _{\left( {{\mathbf{x}},y} \right) \sim \widehat{\mathcal{D}}_t }}\left[ {\theta \left( {\mathbf{x}} \right) \ne y} \right]$, where $\theta\left(\mathbf{x}\right)$ is the task-specific classifier. In our setting, only the source domain is labeled. The bridge and target domains are unlabeled.

\begin{figure*}[t]
    \centering
    \includegraphics[width=\linewidth]{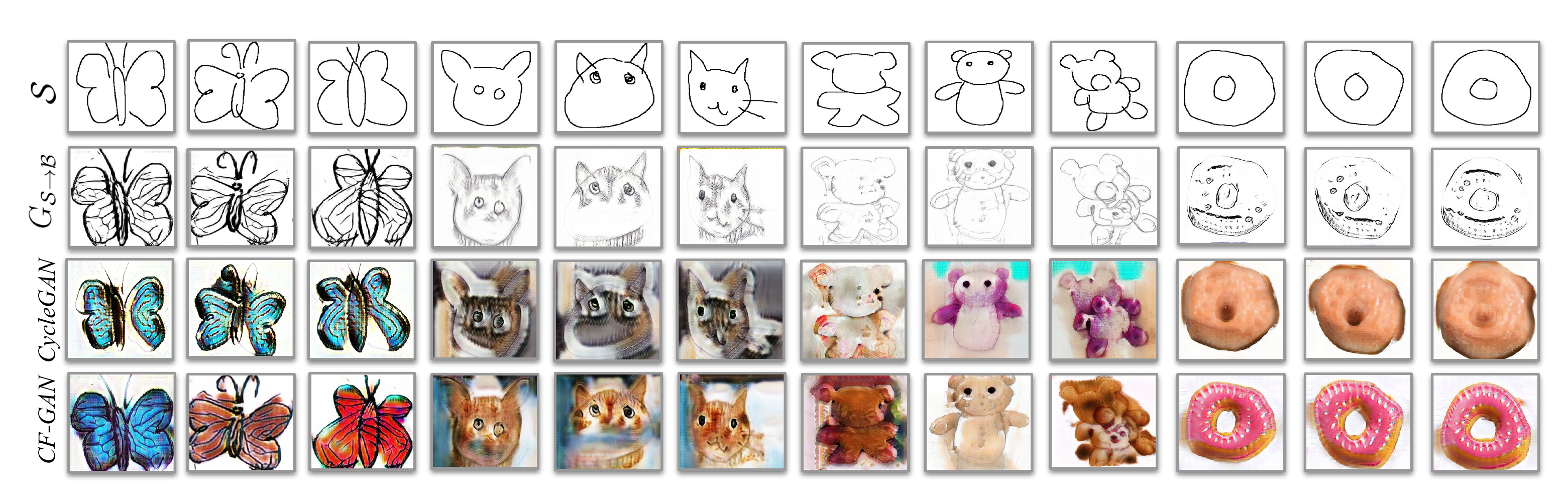}
    \caption{{\bf CFGAN Results.} We take the \textit{Quickdraw} domain, \textit{Sketch} domain, and \textit{Real} domain from DomainNet~\cite{lsdac} as the $\mathcal{D}_s$, $\mathcal{D}_b$, and $\mathcal{D}_t$, respectively. The first row shows the images samples from the source \textit{Quickdraw} domain. The second row shows the output of $G_{\mathcal{S}\rightarrow\mathcal{B}}$. The third and forth row shows the results of CycleGAN~\cite{CycleGAN2017} and our model CFGAN, respectively. The experimental results demonstrate that our model can generate more realistic textures than CycleGAN when the domain shift between the $\mathcal{D}_s$ and $\mathcal{D}_t$ is significantly large.}  

    \label{fig_cf_gan_result}
\end{figure*}

We propose a Prototypical Adversarial Domain Adaptation framework to align the $\mathcal{D}_s$ and $\mathcal{D}_t$, using the intermediate bridge domain $\mathcal{D}_b$. Figure~\ref{fig_PADA_overview} shows the proposed model. The entire framework comprises three components: (\textbf{1}) adversarial domain alignment component (ADA), (\textbf{2}) prototypical matching network component, and (\textbf{3}) disentanglement component. ADA aligns the ($\mathcal{D}_s$, $\mathcal{D}_b$) and ($\mathcal{D}_b$, $\mathcal{D}_t$) pairs with adversarial training process. The prototypical network component is devised to align the support of each domain with Maximum Mean Discrepancy~\cite{gretton2007kernel} (MMD) loss. The disentanglement component is designed to untangle the extracted features to \textit{domain-invariant} ($f_{di}$) and \textit{domain-specific} ($f_{ds}$) features so to limit potential negative transfer.  

Since only the source domain is labeled, we train the feature generator with the following cross-entropy task loss:
\begin{equation}
    \mathcal{L}_{ce} = -\mathbb{E}_{(\mathbf{x}_s,y_s)\sim\widehat{\mathcal{D}}_s} \sum_{k=1}^{K}\mathds{1} [k=y_s]log(C(f_G)),
    \label{equ_cross_entropy}
\end{equation}
where $f_G$ denotes the feature extracted by feature extractor $G$ and $C$ represents the classifier.

\noindent \textbf{Adversarial Alignment} To align the $\mathcal{D}_s$ with $\mathcal{D}_t$, we leverage adversarial alignment approach to first align the ($\mathcal{D}_s$, $\mathcal{D}_b$) and then the ($\mathcal{D}_b$,$\mathcal{D}_t$) pairs. This is achieved by exploiting adversarial domain classification in the resulting latent space. Specifically, we leverage a domain identifier $DI$, which takes the disentangled feature ($f_{di}$ or $f_{ds}$ ) as input and outputs the domain label $l_f$ (source or target). The objective function of the domain identifier is as follows:
\begin{equation}
\label{eqn_domain_identifier}
\mathcal{L}_{DI} = - \mathbb{E}[l_f\log P( l_f)]  + \mathbb{E}(1-l_f)[\log P(1-l_f )],
\end{equation}
Then the feature extractor $G$ is trained to fool the domain identifier $DI$ so that $DI$ can not recognize which domain a given feature vector belongs to.

\noindent \textbf{Prototypical Network Matching} Prototypical networks calculate an $M$-dimensional representation $\mathbf{c}_k \in \mathbb{R}^M$, or \emph{prototype}, of each class through an embedding function $f_{\bm \phi} : \mathbb{R}^D \rightarrow \mathbb{R}^M$ with learnable parameters $\bm{\phi}$. Empirically, Each prototype is the mean vector of the embedded support points belonging to its class:
\begin{equation}
    \mathbf{c}_k = \frac{1}{|S_k|} \sum_{(\mathbf{x}_i, y_i) \in S_k} f_{\bm{\phi}}(\mathbf{x}_i)
   \label{eq:prototype}
\end{equation}
Given a distance function $d: \mathbb{R}^M \times \mathbb{R}^M \rightarrow [0, +\infty)$, prototypical networks produce a distribution over classes for a query point $\mathbf{x}$ based on a softmax over distances to the prototypes in the embedding space:
\begin{equation}
    p_{\bm \phi}(y = k\,|\,\mathbf{x}) = \frac{\exp(-d(f_{\bm \phi}(\mathbf{x}), \mathbf{c}_k))}{\sum_{k'} \exp(-d(f_{\bm \phi}(\mathbf{x}), \mathbf{c}_{k'}))}
    \label{eq:classdist}
\end{equation}
Learning proceeds by minimizing the negative log-probability $J(\bm{\phi}) = -\log p_{\bm \phi}(y = k \,|\,\mathbf{x})$ of the true class $k$ via SGD. Training episodes are formed by randomly selecting a subset of classes from the training set, then choosing a subset of examples within each class as the support set and the rest serving as query points.

Conventional domain adaptation methods~\cite{long2015,JAN,adda,DANN} only align the marginal probability of the extracted features, while many state-of-the-art works~\cite{gong2016domain, MCD_2018, pan2019transferrable} argue that only aligning the marginal probability achieve limited domain adaptation performance. Recently, Pan \textit{et al}~\cite{pan2019transferrable} proposes to align the class-level distribution by computing the pairwise distance between the prototypes of the same class from different domains in reproducing kernel Hilbert space. The basic intuition is that if the data distributions of \textit{source} and \textit{target} domains are identical, the prototypes of the same class in these domains are also the same. The class-level discrepancy loss is defined as follows:
\begin{equation}\label{Eq:Eq5}
\begin{array}{l}
{L_G}\left( {\left\{ {\mu _{_c}^s} \right\},\left\{ {\mu_{_c}^t} \right\},\left\{ {\mu _{_c}^{b}} \right\}} \right) {\buildrel \Delta \over =} \frac{1}{C}\sum\limits_{c = 1}^C {\left\| {\widetilde\mu _{_c}^s - \widetilde\mu _{_c}^t} \right\|_{\mathcal H}^2}\\
~~~~~~~~+ \frac{1}{C}\sum\limits_{c = 1}^C {\left\| {\widetilde\mu _{_c}^s - \widetilde\mu _{_c}^{b}} \right\|_{\mathcal H}^2} + \frac{1}{C}\sum\limits_{c = 1}^C {\left\| {\widetilde\mu _{_c}^t - \widetilde\mu _{_c}^{b}} \right\|_{\mathcal H}^2},
\end{array}
\end{equation}
where $\left\{ {\widetilde\mu _{_c}^s} \right\}$, $\left\{ {\widetilde\mu _{_c}^t} \right\}$ and $\left\{ {\widetilde\mu _{_c}^{b}} \right\}$ denote the corresponding prototypes of source, target, and bridge domains in reproducing kernel Hilbert space ${\mathcal H}$. Minimizing this term results in decreasing the distance between the prototype of each class computed in each domain. We utilize the bridge domain to better align the \textit{source} and the \textit{target} domain in the embedding space. 

\begin{figure}[t]
    \centering
    \includegraphics[width=\linewidth]{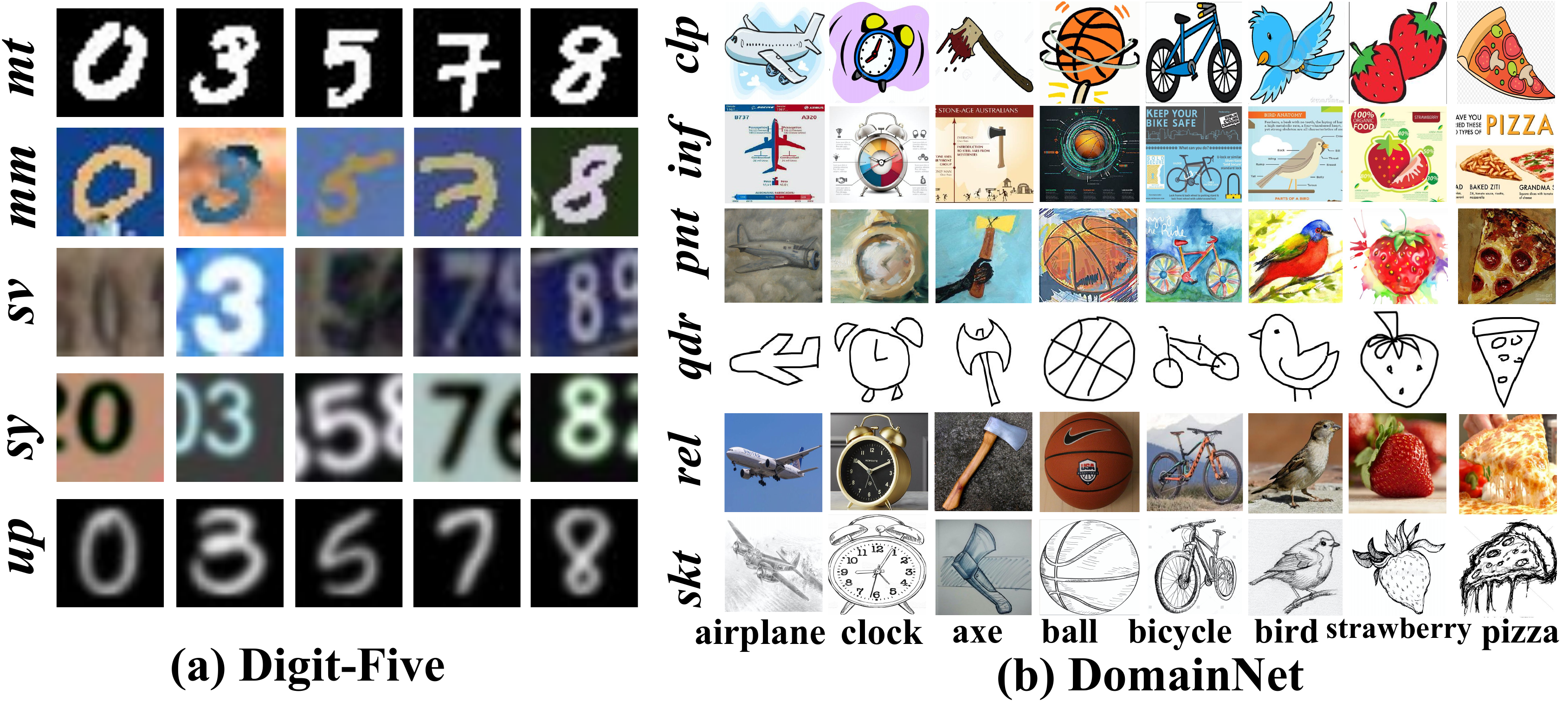}
    \caption{Two benchmarks we use in our paper.}

    \label{fig_dataset_overview}
    \vspace{-0.4cm}
\end{figure}

\noindent \textbf{Disentanglement} To address the above problem, we employ \textit{class disentanglement} to remove class-irrelevant features, such as background, in an adversarial manner.
First, we train a disentangler $D$ and the $K$-way class identifier $C$ to improve label prediction,
supervised by the cross-entropy loss:
\begin{equation}
    \mathcal{L}_{ce} = -\mathbb{E}_{(x_s,y_s)\sim\widehat{\mathcal{D}}_s} \sum_{k=1}^{K}\mathds{1} [k=y_s]log(C(f_D))
    \label{equ_cross_entropy}
\end{equation}
where $f_D\in\{f_{di}, f_{ci}\}$.

In the second step, we fix the class identifier $C$ and train the disentangler $D$ to fool $C$ by generating class-irrelevant features $f_{ci}$. This can be achieved by minimizing the negative entropy loss of the 
predicted class distribution:
\begin{equation}
\mathcal{L}_{ent} = - \frac{1}{n_s} \sum_{j=1}^{n_s} \log C(f^j_{ci}) - \frac{1}{n_t} \sum_{j=1}^{n_t} \log C(f^j_{ci})
\label{equ_entropy_loss}
\end{equation}
where the first and second term indicate entropy loss minimization on the source and the heterogeneous target domain, respectively. 
The above adversarial training process forces the corresponding \textit{disentangler} to extract \textit{class-irrelevant} features. 

\noindent \textbf{Mutual Information Minimization} To enhance the disentanglement, we minimize the mutual information shared between \textit{domain-invariant} features and \textit{domain-specific} features, following Peng \textit{et al}~\cite{DAL_DADA}. Specifically, the mutual information is defined as $I({f}_{di}; {f}_{ds}) = \int_{\mathcal{P} \times \mathcal{Q}} \log{\frac{d\PJ{P}{Q}}{d\PI{P}{Q}}} d\PJ{P}{Q}$, where $\PJ{P}{Q}$ is the joint probability distribution of (${f}_{di}, {f}_{ds}$), and $\mathbb{P}_{\mathcal{P}} =
\int_{\mathcal{Q}} d\PJ{P}{Q}$, $\mathbb{P}_{\mathcal{Q}} = \int_{\mathcal{Q}} d\PJ{P}{Q}$ are the marginals. Despite being a pivotal measure across different distributions, the mutual information is only tractable for discrete variables in cases where the probability distributions are unknown~\cite{mine}. Following Peng \textit{et al}, we adopt the Mutual Information Neural Estimator (MINE)~\cite{mine} to estimate mutual information by using a neural network $T_{\theta}$: 
$\widehat{I(\mathcal{P};\mathcal{Q})}_n = \sup_{\theta \in \Theta} \mathbb{E}_{\mathbb{P}^{(n)}_{\mathcal{P}\mathcal{Q}}}[T_\theta] - \log(\mathbb{E}_{\mathbb{P}^{(n)}_{\mathcal{P}} \otimes \widehat{\mathbb{P}}^{(n)}_{\mathcal{Q}}}[e^{T_\theta}])$. Practically, MINE can be calculated as $I(\mathcal{P};\mathcal{Q})=\int\int{\mathbb{P}^{n}_{\mathcal{P}\mathcal{Q}}(p,q)\text{ }T(p,q,\theta)}$  -  $\log (\int\int\mathbb{P}^{n}_{\mathcal{P}}(p)\mathbb{P}^n_{\mathcal{Q}}(q)e^{T(p,q,\theta)})$. To avoid computing the integrals, we use Monte-Carlo integration to calculate the estimation:
\begin{equation}
I(\mathcal{P},\mathcal{Q})=\frac{1}{n}\sum^{n}_{i=1}T(p,q,\theta)-\log(\frac{1}{n}\sum_{i=1}^{n}e^{T(p,q',\theta)})
\label{equ_mutual_information}
\end{equation}
where $(p,q)$ are sampled from the joint distribution of (${f}_{di}, {f}_{ds}$) and $q'$ is sampled from the marginal distribution. The \textit{domain-invariant} and \textit{domain-specific} features are forwarded to a reconstructor with a L2 loss to reconstruct the original features so to maintain the representation integrity, as shown in Figure~\ref{fig_PADA_overview}.

\section{Experiments}

\begin{table}[t]
\centering
\scalebox{0.85} {
\begin{tabular}{lcccc}
 & & & \\
\textbf{Method} & \textbf{{\BV{qdr}$\rightarrow$\BV{rel}} acc.} & \textbf{{\BV{rel}$\rightarrow$\BV{qdr}} acc.} & \textbf{Mean acc.} \\ \hline
Source only & 0.31 & 0.13 & 0.22 \\
CycleGAN & 0.35 & 0.15 & 0.25 \\
ADDA& 0.46 & 0.20 & 0.33 \\ 
MCD&0.49 & 0.17 & 0.33\\
DAN&  0.50 & 0.18 & 0.34 \\ 
SE & 0.48 & 0.22& 0.35 \\
\textbf{CFGAN} (ours) & \textbf{0.52} & \textbf{0.27} & \textbf{0.39} \\  
\end{tabular} }
\caption{\rebuttal{Cross-domain recognition results. Evaluated on 10 categories selected from DomainNet. }}
\label{cfgan_cross_domain}
\vspace{-0.4cm}
\end{table}

Our experiments include two parts. First, we apply CFGAN to translate quickdraw images provided by ~\cite{lsdac} to real images. Second, we apply the PADA framework to unsupervised domain adaptation and test our model on the DomainNet~\cite{lsdac} benchmark. We implement our model with PyTorch and train it on a clusters with ten Nvidia TitanX GPUs. Datasets, code, and experimental configurations will be made available publicly. 

\subsection{Image-Image Generation with CFGAN}
\label{exp_i2i_generation}

\begin{table*}[t]

    \addtolength{\tabcolsep}{1pt}
    \centering
    
    \label{table:office10}
    \begin{tabular}{cccccc}
        \Xhline{1pt}
        Method & \footnotesize{SVHN $\xrightarrow{\text{\BV{mm}}}$ MNIST}  & \footnotesize{MNIST $\xrightarrow{\text{\BV{mm}}}$ SVHN} &  \footnotesize{USPS $\xrightarrow{\text{\BV{sy}}}$ SVHN} & \footnotesize{SVHN $\xrightarrow{\text{\BV{sy}}}$ USPS}  & Average \\
        \hline
        Source Only   & 63.4$\pm$1.8 & 11.7$\pm$0.7 & 13.5$\pm$0.5 & 75.9$\pm$1.3 & 41.1$\pm$1.1\\
          DANN~\cite{DANN}   & {\textcolor{red}{65.4}} (67.3$\pm$1.4)& {\rebuttal{17.7}} ({19.8$\pm$0.9}) & \textcolor{blue}{20.4} (20.2$\pm$0.7)&\textcolor{red}{75.4} (76.6$\pm$1.1) & \textcolor{red}{44.7}(46.0$\pm$1.0)\\
        DAN~\cite{long2015} &\textcolor{blue}{70.4} (68.7$\pm$1.4)&\textcolor{blue}{21.5} (20.4$\pm$0.8) & \textcolor{blue}{22.9} (22.3$\pm$0.6) & \textcolor{blue}{80.2} \textbf({79.0}$\pm$1.4) & \textcolor{blue}{48.8} (47.6$\pm$0.9)  \\
        ADDA~\cite{adda} &\textcolor{red}{69.2} (70.1$\pm$1.5) & \textcolor{red}{19.7} (21.5$\pm$0.7) & \textcolor{blue}{24.2} (23.4$\pm$0.8) & \textcolor{red}{76.9} (78.4$\pm$1.2) &\textcolor{red}{47.5} (48.4$\pm$1.2) \\
      
        \textbf{PADA} (Ours) & \textbf{72.1}$\pm$1.3 & \textbf{24.3}$\pm$0.8 & \textbf{25.2}$\pm$0.9 & 78.5$\pm$1.2 & \textbf{50.1}$\pm$1.1\\
        
	     \hline
        
        \Xhline{1pt}
    \end{tabular}
    \caption{Accuracy on ``Digit-Five'' dataset with large domain gap learning protocol. Our model PADA achieves \textbf{50.1}\% accuracy, outperforming other baselines. {\BV{mm}},{\BV{sy}}  are abbreviations for \textit{MNIST-M}~\cite{DANN}, \textit{Synthetic Digits}~\cite{DANN}. {\rebuttal{We run the baselines with the blending-target~\cite{chen2019blending}} schema, and find that the performance of discrepancy-alignment method (DAN) increases (denoted by \textcolor{blue}{blue}) and the performance of adversarial-alignment methods (ADDA, DANN) decrease (denoted by \textcolor{red}{red}).}}
    \label{table_digit_five}
\end{table*}

\begin{figure*}[t]
\vspace{-0.3cm}
    \begin{minipage}{\hsize}
      \centering
      \subfigure[\scriptsize Source Features ]
      {\includegraphics[width=0.215\hsize]{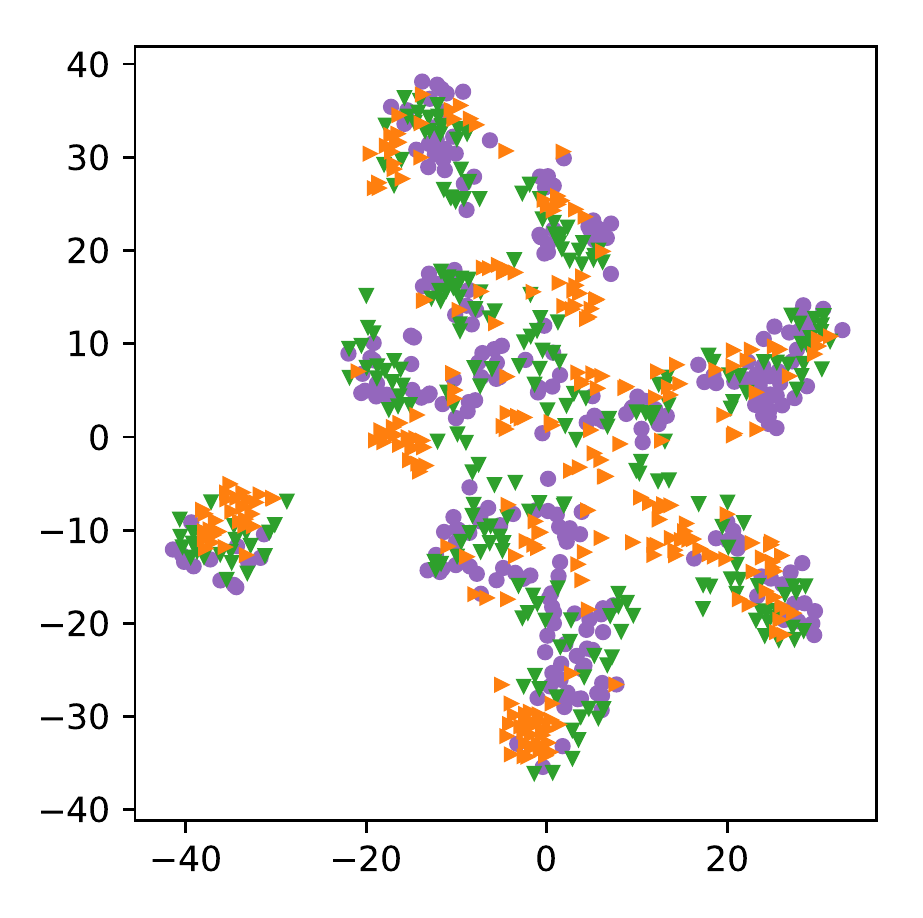}
      \label{tsne_source} }
     \centering
      \subfigure[\scriptsize DAN Features ]
      {\includegraphics[width=0.215\hsize]{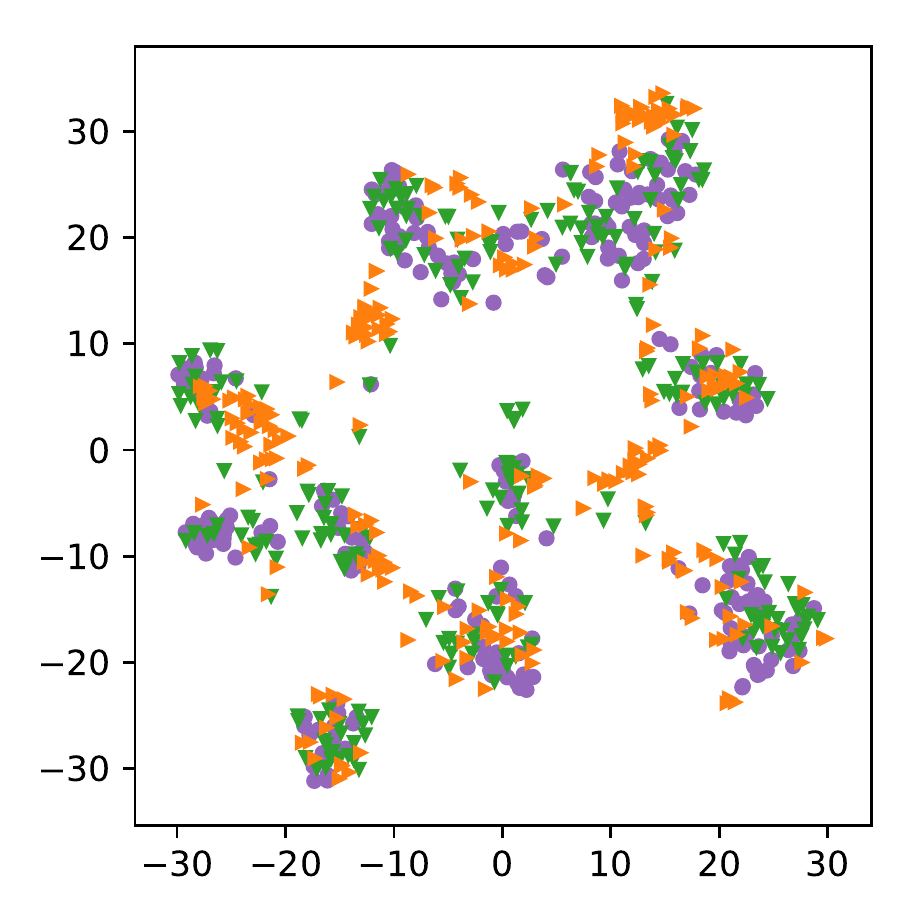}
      \label{tsne_fdann}}
      \centering
      \subfigure[\scriptsize ADDA Features]
      {\includegraphics[width=0.215\hsize]{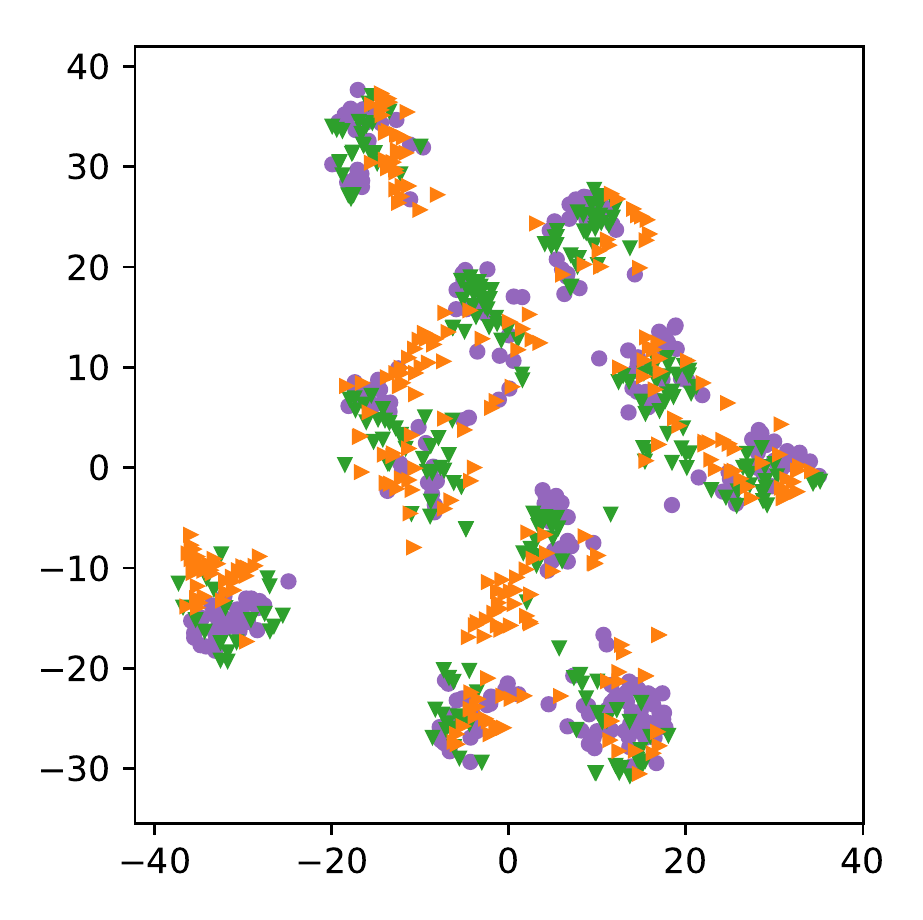} 
      \label{tsne_fdan}}
      \centering
      \subfigure[\scriptsize PADA Features ]
      { \includegraphics[width=0.215\hsize]{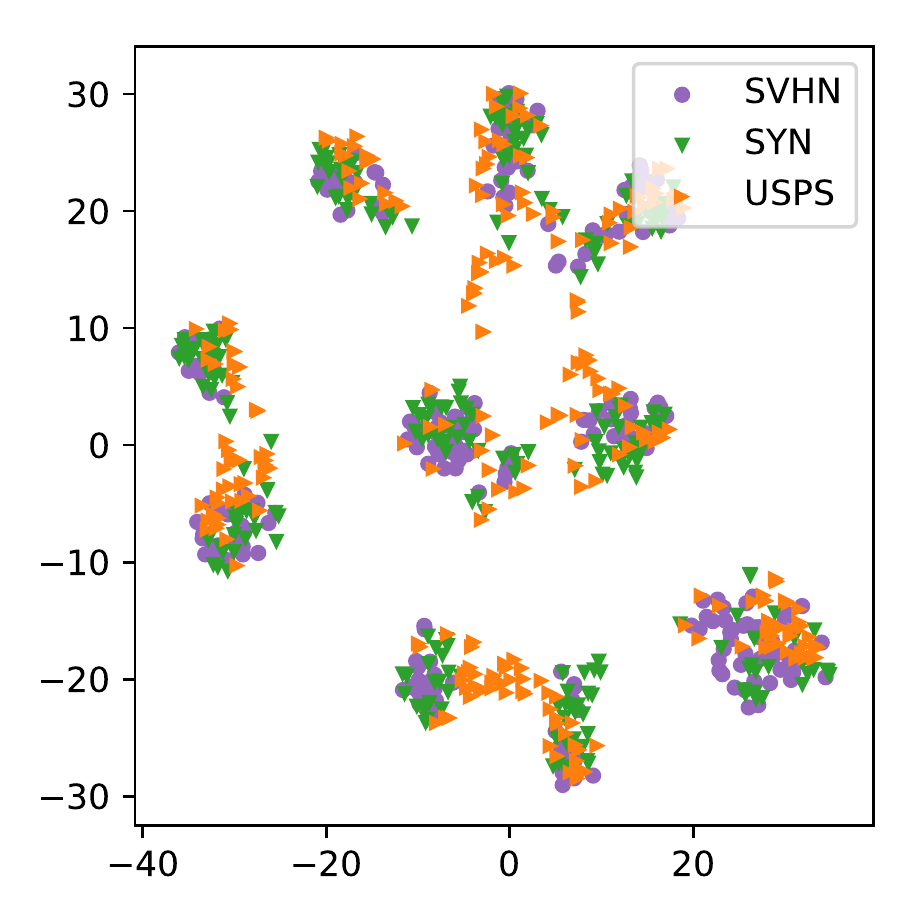}
      \label{tsne_pada}}
    \end{minipage}
  \caption{\small Feature visualization: t-SNE plot of source-only features, DAN~\cite{long2015} features, ADDA~\cite{adda} features, and PADA features in \textbf{SVHN} $\rightarrow$ \textbf{USPS} setting. We use different markers and colors to denote different domains. (Best viewed in color.)}
  \label{fig_analysis}
  \vspace{-0.3cm}
\end{figure*}

\noindent \textbf{DomainNet} The DomainNet dataset contains six distinct domains and about 0.6 million images distributed among 345 categories. It comprises six domains: \textit{Clipart} ({\BV{clp}}), a collection of clipart images; \textit{Infograph} ({\BV{inf}}), infographic images with specific object; \textit{Painting}, artistic depictions of object in the form of paintings; \textit{Quickdraw} ({\BV{qdr}}), drawings from the worldwide players of game ``Quick Draw!"\footnote{\url{https://quickdraw.withgoogle.com/data}}; \textit{Real} ({\BV{rel}}), photos and real world images; and \textit{Sketch} ({\BV{skt}}), sketches of specific objects. DomainNet is by far the largest dataset collected specifically for domain adaptation tasks, containing informative vision cues across domains.

For our qualitative experiment, we select the \textit{Quickdraw} domain as $\mathcal{D}_s$, \textit{Real} domain as $\mathcal{D}_t$, and \textit{Sketch} domain as $\mathcal{D}_b$, respectively. The network architectures for ($G_{\mathcal{S}\rightarrow\mathcal{B}}$, $G_{\mathcal{B}\rightarrow\mathcal{T}}$, $F_{\mathcal{S}\rightarrow\mathcal{B}}$,
$F_{\mathcal{B}\rightarrow\mathcal{T}}$) and ($D_{\mathcal{S}}$, $D_{\mathcal{B}}$, $D_{\mathcal{T}}$) are identical. 

Figure~\ref{fig_cf_gan_result} compares the experimental results between CycleGAN and our CFGAN. We can make the following observations: (\textbf{1}) our model CFGAN can render more realistic images than CycleGAN when the domain shift between $\mathcal{D}_s$ and $\mathcal{D}_t$ is significantly large. (\textbf{2}) the outputs of $G_{\mathcal{S}\rightarrow\mathcal{B}}$ show that the bridge domain preserves local texture in transferring from \textit{Quickdraw} to \textit{Real} domain. For example, CFGAN can generate realistic textures on the wings of butterflies and render realistic fur on the head of cats. 

{\rebuttal{To demonstrate the effectiveness of our proposed CFGAN on the cross-domain image recognition task, we compare CFGAN with state-of-the-art domain adaptation approaches on 10 categories selected from DomainNet~\cite{lsdac} (ballon, butterfly, cat, dog, donut, horse, grapes, pineapple, sheep, teddy bear). We compute the mean accuracy on {\BV{qdr}$\rightarrow$\BV{rel}} and {\BV{rel}$\rightarrow$\BV{qdr}} settings. As Table~\ref{cfgan_cross_domain} shows, our model can improve the performance of cross-domain recognition by a large margin.}}

\subsection{Experiments On Digits Datasets}
In this experiment, we evaluate our PADA model under two different translation schemes. The data samples are presented in Figure~\ref{fig_dataset_overview}.

\noindent \textbf{Digit-Five} 
We also conduct experiments on five digit datasets, namely MNIST ({\BV{mt}})~\cite{lecun1998gradient}, SVHN ({\BV{sv}})~\cite{svhn}, MNIST-M ({\BV{mm}})~\cite{DANN}, Synthetic Digits ({\BV{sy}})~\cite{DANN}, USPS~\cite{usps} ({\BV{up}}). 
 
 \begin{table*}[t]

    \addtolength{\tabcolsep}{0.01pt}
 
    \vspace{-0.2cm}
    \centering
    
    \label{table:office10}
    \begin{tabular}{cccccc}
        \Xhline{1pt}
        Method &\footnotesize{Real $\xrightarrow{\text{\BV{clp}}}$ Sketch} &  \footnotesize{Quickdraw $\xrightarrow{\text{\BV{skt}}}$ Real}  &  \footnotesize{Quickdraw $\xrightarrow{\text{\BV{skt}}}$ Clipart}  & \footnotesize{Quickdraw $\xrightarrow{\text{\BV{skt}}}$ Infograph} &  Average \\
        \hline
        Source Only   & 23.1$\pm$0.5 & 5.5$\pm$0.2 &13.4$\pm$0.5 &  1.2$\pm$0.1 &  10.8$\pm$0.3 \\
        DAN~\cite{long2015} & \textcolor{blue}{28.3} (\textbf{26.2}$\pm$0.7) & \textcolor{blue}{10.2} (8.5$\pm$0.3) & \textcolor{blue}{14.9} (14.2$\pm$0.5) & 1.6 (1.6$\pm$0.1) &\textcolor{blue}{13.8} (12.6$\pm$0.4)  \\
        DANN~\cite{DANN}   & \textcolor{red}{22.4} (24.5$\pm$0.6) &\textcolor{red}{6.5} (8.7$\pm$0.4)&\textcolor{red}{15.3} (16.8$\pm$0.4) &  \textcolor{blue}{1.9} (1.8$\pm$0.1)&  \textcolor{red}{11.5} (12.9$\pm$0.4) \\
        ADDA~\cite{adda} &\textcolor{red}{23.0} (25.5$\pm$0.3) &\textcolor{red}{8.5} (9.1$\pm$0.5) &\textcolor{red}{14.9} (15.7$\pm$0.3) & 2.3 (2.3$\pm$0.2) & \textcolor{red}{12.2} (13.2$\pm$0.3) \\
        \textcolor{black}{SE~\cite{SE}} &20.7$\pm$0.4&6.4$\pm$0.5&12.9$\pm$0.5&1.5$\pm$0.2&10.3$\pm$0.4 \\
        \textcolor{black}{MCD~\cite{MCD_2018}} &25.4$\pm$0.5 & 8.2$\pm$0.4 & 14.8$\pm$0.4 & 2.1$\pm$0.2 & 12.6$\pm$0.4\\
        \textbf{PADA} (Ours) &\textbf{26.2}$\pm$0.3 & \textbf{9.7}$\pm$0.4 & \textbf{17.4}$\pm$0.6 & \textbf{3.2}$\pm$0.3 &  \textbf{14.2}$\pm$0.4 \\
        
	     \hline
        
        \Xhline{1pt}
    \end{tabular}
    \vspace{0.2cm}
    \caption{Accuracy on DomainNet~\cite{lsdac} dataset with large domain gap learning protocol. Our model PADA achieves \textbf{14.2}\% accuracy, outperforming other baselines. {\BV{clp}}, {\BV{skt}}  are abbreviations for \textit{Clipart}, \textit{Sketch}.}
    \label{table_domainnet}
\end{table*}

 \noindent $\textbf{SVHN}\xrightleftharpoons[]{\text{\BV{mm}}}\textbf{MNIST}$ The domain discrepancy between SVHN and MNIST datasets is very large as SVHN dataset often contains images with colored background, multiple digits, as well as blurry digits. We employ MNIST-M ({\BV{mm}}) dataset as the bridge domain because digit images from the ({\BV{mm}}, {\BV{sv}}) domain pair are both of colored background, meanwhile digits from the ({\BV{mm}}, {\BV{mt}}) domain pair are in similar calligraphic style. 

  \noindent $\textbf{USPS}\xrightleftharpoons[]{\text{\BV{sy}}}\textbf{SVHN}$ The domain shift between USPS and SVHN is significantly large as SVHN dataset contains images with colored background, while the images in USPS dataset are only composed of black background. In this experiment, we utilize Synthetic Digits dataset ({\BV{sy}}) as the bridge domain. Similar to ({\BV{sv}}) dataset, images in ({\BV{sy}}) dataset constitutes colored background and multiple digits.

\begin{table*}[t]
\vspace{-0.2cm}
\small
\begin{center}
\begin{tabular}
{ p{0.7cm} p{1.4cm} p{1.4cm} p{1.4cm} p{1.4cm} p{1.4cm} p{1.4cm} p{1.4cm} p{1.4cm} p{1.4cm} p{1.4cm} p{1.4cm} }

&\includegraphics[width=0.99\linewidth]{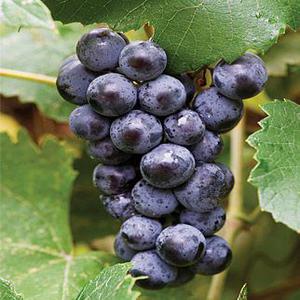}
&\includegraphics[width=0.99\linewidth]{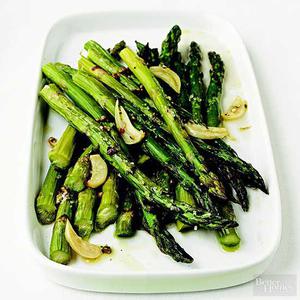}
&\includegraphics[width=0.99\linewidth]{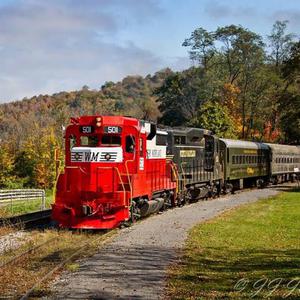}
&\includegraphics[width=0.99\linewidth]{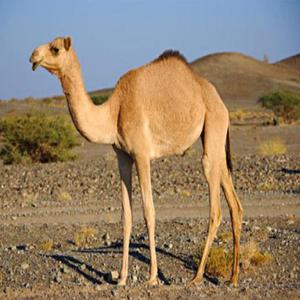}
&\includegraphics[width=0.99\linewidth]{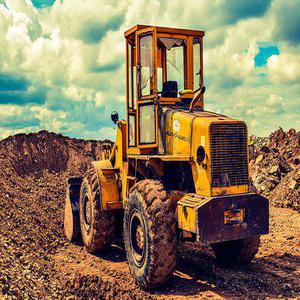}
&\includegraphics[width=0.99\linewidth]{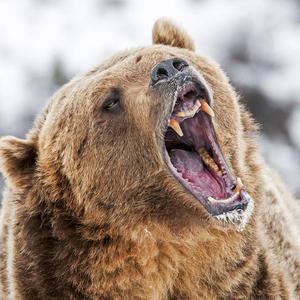}
&\includegraphics[width=0.99\linewidth]{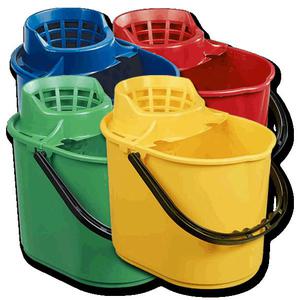}
&\includegraphics[width=0.99\linewidth]{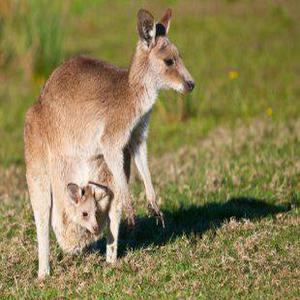}
&\includegraphics[width=0.99\linewidth]{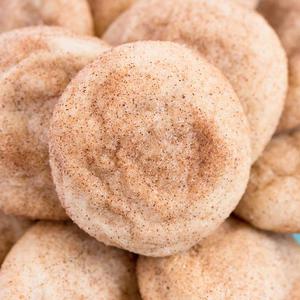}\\
\hline
\textbf{GT} &bear&grape&train&camel&bulldozer&bear&bucket&kangaroo&cookie\\

\textbf{DANN} &\textcolor{red}{bluberry} & \textcolor{red}{broccoli}& \textcolor{red}{bus} &\textcolor{red}{horse} &\textcolor{red}{truck} &\textcolor{red}{lion} & \textcolor{red}{basket} & \textcolor{ForestGreen}{kangaroo} & \textcolor{red} {bread} \\

\textbf{Ours} & \textcolor{ForestGreen}{bear} &\textcolor{ForestGreen}{grape} &\textcolor{ForestGreen}{train} &\textcolor{ForestGreen}{camel} &\textcolor{ForestGreen}{bulldozer} &\textcolor{ForestGreen}{bear} &\textcolor{ForestGreen}{bucket} &\textcolor{red}{sheep} &\textcolor{red}{cake} 
\end{tabular}
\end{center} 
\vspace{-0.2cm}
\caption{Prediction examples of DANN and our model on \textit{Quickdraw}$\rightarrow$\textit{Real} setting. We show examples where our model improves on the baseline, as well as typical failure cases.}
\label{tab_cad_dgcan}
\vspace{-0.2in}
\end{table*}

 We compare our model to state-of-the-art baselines: Deep Adaptation Network (\textbf{DAN})~\cite{long2015}, Domain Adversarial Neural Network (\textbf{DANN})~\cite{DANN}, Adversarial Discriminative Domain Adaptation (\textbf{ADDA})~\cite{adda}. 
 {\rebuttal{For fair comparison, we combine the bridge domain and target domain to a blending-target domain}, following the schema proposed by~\cite{chen2019blending}}
 
 \noindent \textbf{Results} The experimental results on the ``Digit-Five'' dataset are shown in Table~\ref{table_digit_five}. We can observe: (1) our model achieves an average accuracy of \textbf{50.1}\% , outperforming almost all other baselines in the large-shift domain adaptation tasks. (2) PADA improves the performance of the scarcely studied setting where SVHN dataset is selected as the target domain, demonstrating the effectiveness of our approach.
 
 To dive deeper into the PADA features, we plot the t-SNE embeddings of the feature representations generated by the source only model, DAN, ADDA, and PADA on the {\footnotesize{SVHN $\xrightarrow{\text{\BV{sy}}}$ USPS}} task in Figure~\ref{tsne_source}-\ref{tsne_pada}. We observe that the features 
 extracted by our model are more well separated between classes than DAN and ADDA features.

 \subsection{Experiments on DomainNet}
 To demonstrate the effectiveness of our model on object-level image recognition tasks, we conduct experiments on DomainNet~\cite{lsdac}. 
 The data samples are presented in Figure~\ref{fig_dataset_overview}.
 
 \noindent\textbf{Real}$\xrightarrow{\text{\BV{clp}}}$\textbf{Sketch} While images from domain pair (\textit{Real}, \textit{Clipart}) are similar in color patterns, the ones from the domain pair (\textit{Clipart}, \textit{Sketch}) are of similar stroke style.

 \noindent\textbf{Quickdraw}$\xrightarrow{\text{\BV{skt}}}$\textbf{Real}/\textbf{Clipart}/\textbf{Infograph} The domain discrepancy between \textit{Quickdraw} and \textit{Real} domain is significantly large. The images from \textit{Real} domain contain rich visual cues such as color, texture, and background, while images in the \textit{Quickdraw} domain are solely composed of simple strokes. In this experiment, we use the \textit{Sketch} ({\BV{skt}}) domain as the bridge domain.

\noindent\textbf{Results} The experimental results on DomainNet are shown in Table~\ref{table_domainnet}. Our model achieves \textbf{14.2}\% accuracy and outperforms all other baselines, demonstrating the effectiveness of our model tackling large domain shift on large-scale dataset. Note that this datasaet contains 0.6 million images and so even a one-percent performance improvement is not trivial. 

\begin{figure}[t]
    \vspace{-0.5cm}
    \begin{minipage}{\hsize}
      \centering
      \subfigure[\scriptsize $\mathcal{A}$-Distance ]
      {\includegraphics[width=0.42\hsize]{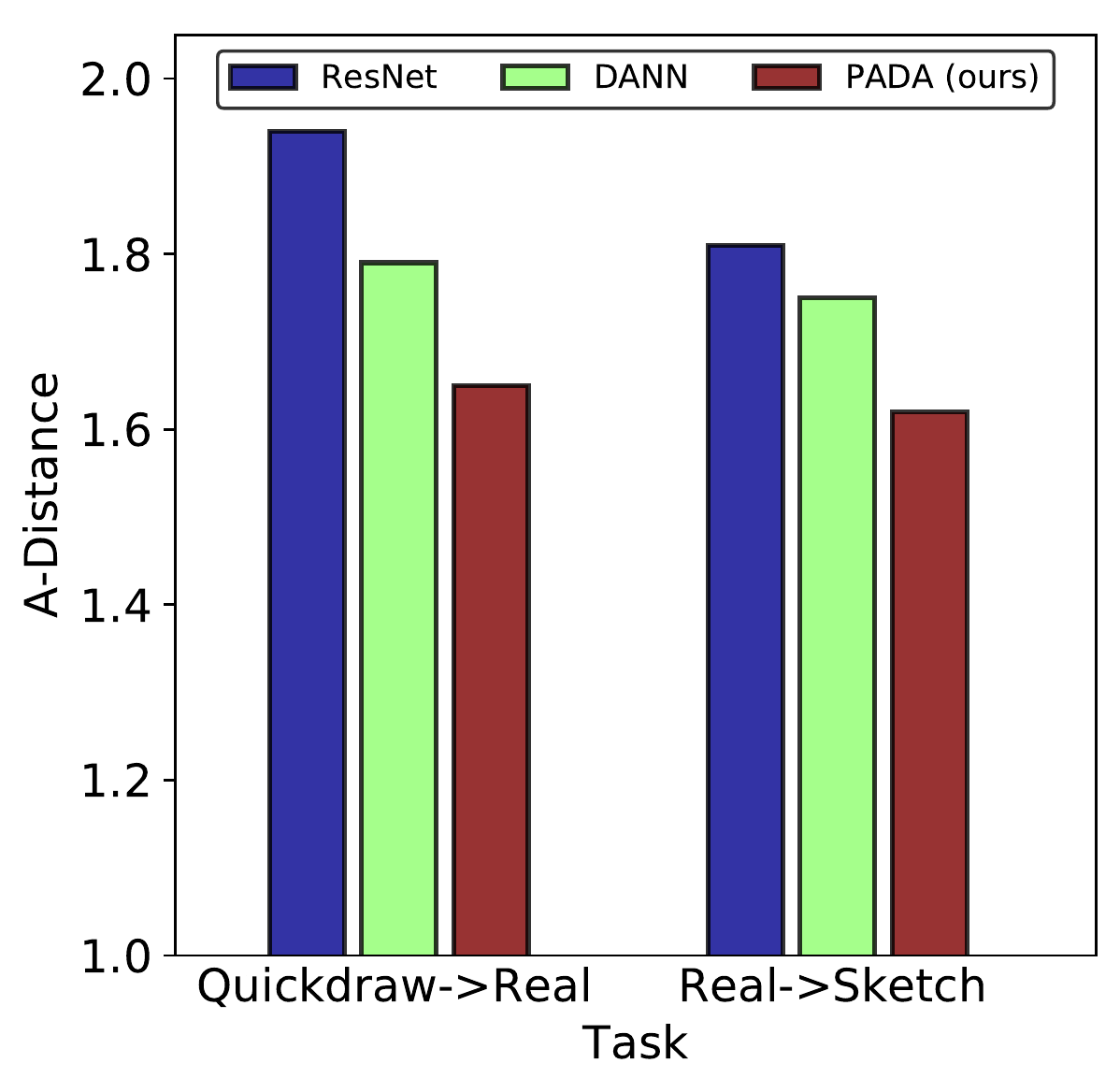}
      \label{a_distance} }
     \centering
      \subfigure[\scriptsize Training loss for Real$\rightarrow$Sketch]
      {\includegraphics[width=0.53\hsize]{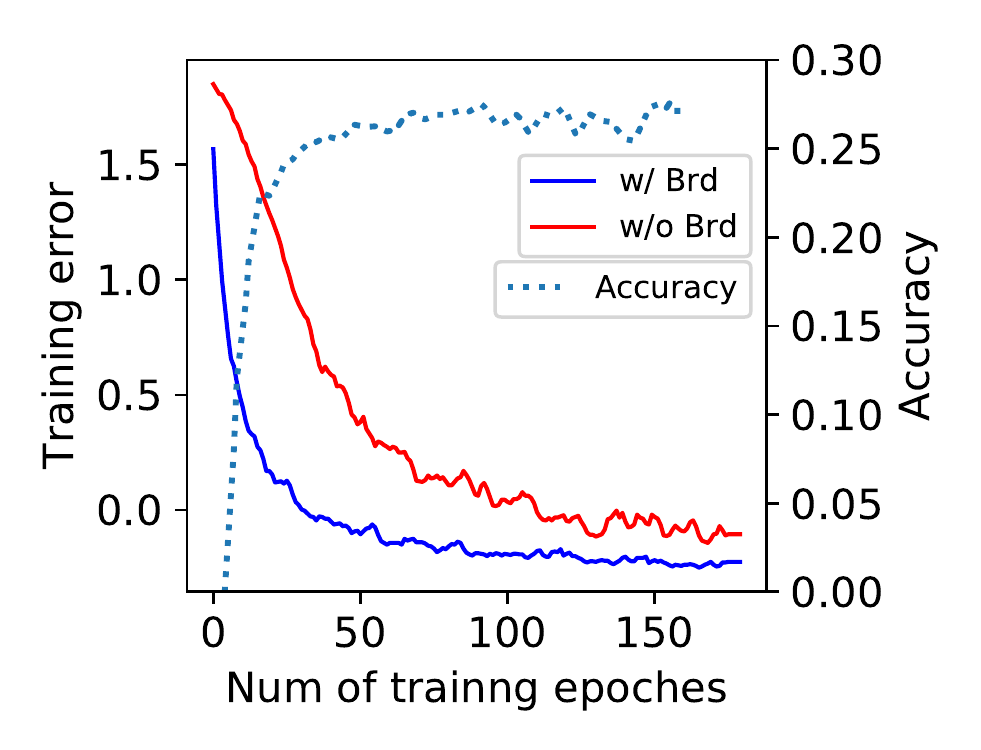}
      \label{domainnet_loss}}
      
    \end{minipage}
    \vspace{-0.2cm}
  \caption{\small(\textbf{a})$\mathcal{A}$-Distance of ResNet, DANN, and FADA features on two different tasks. (\textbf{b}) training errors and accuracy on Real$\rightarrow$Sketch task.}
  
  \label{fig_domainnet_loss}
  \vspace{-0.5cm}
\end{figure}

To better analyze the effectiveness of PADA, we perform the following analyses: \textbf{(1)} \textbf{$\mathcal{A}$-distance} Ben-David \textit{et al}~\cite{ben2010theory} proposes $\mathcal{A}$-distance to evaluate the domain discrepancy. We calculate  $\mathcal{A}$-distance ${\hat d_{\cal A}} = 2\left( {1 - 2\epsilon } \right)$ for Quickdraw$\rightarrow$Real and Real$\rightarrow$Sketch tasks, where $\epsilon$ is the generalization error of a two-sample classifier (\textit{e.g.} kernel SVM) trained on the binary problem distinguishing input samples as coming from the source or the target domain. We plot ${\hat d_{\cal A}}$ with source-only features, DANN features and PADA features in Figure~\ref{a_distance}. We observe that the ${\hat d_{\cal A}}$ on PADA features is smaller than other baselines, demonstrating that PADA features are harder to be distinguished between source and target. \textbf{(2)} We plot the training error \textit{w/} or \textit{w/o} domain bridge for Real$\rightarrow$Sketch task in Figure~\ref{domainnet_loss}. The figure shows that the training error is smaller when the domain bridge is applied, which is consistent with our quantitative results. \textbf{(3)} We show the predictions of DANN and our model on the \textit{Quickdraw}$\rightarrow$\textit{Real} task. We show examples where our model outperforms the baseline, as well as typical failure cases. 

\section{Conclusion}

In this paper, we first propose a challenging transfer learning paradigm where the target domain is significantly gapped from the source domain. To tackle this task, we have proposed the method of leveraging intermediate domains to bridge knowledge transfer from the source domain to the target. We have presented a generative model called Cycle-consistency Flow Generative Adversarial Networks for image-to-image generation and a deterministic Prototypical Adversarial Domain Adaptation model for unsupervised domain adaptation. Empirically, we demonstrate that our CF-GAN model can generate more realistic images than Cycle-GAN model in the large domain gap scenario. An extensive empirical evaluation of our model on the unsupervised domain adaptation benchmarks demonstrates the efficacy of our proposed PADA model against several state-of-the-art domain adaptation algorithms.

{\small
\bibliographystyle{ieee}
\bibliography{egbib}
}
\clearpage
\section{Model Architecture}

{\bf Generator architectures}
Let \texttt{c7s1-k} denote a $7\times7$ Convolution-InstanceNorm-ReLU layer with $k$ filters and stride $1$. \texttt{dk} denotes a $3\times3$ Convolution-InstanceNorm-ReLU layer with $k$ filters and stride $2$. Reflection padding was used to reduce artifacts. \texttt{Rk} denotes a residual block that contains two $3\times3$ convolutional layers with the same number of filters on both layer. \texttt{uk} denotes a $3\times3$ fractional-strided-Convolution-InstanceNorm-ReLU layer with $k$ filters and stride $\frac{1}{2}$.

The network with 6 residual blocks consists of:\\
\texttt{c7s1-64,d128,d256,R256,R256,R256,\\
R256,R256,R256,u128,u64,c7s1-3}

The network with 9 residual blocks consists of:\\
\texttt{c7s1-64,d128,d256,R256,R256,R256,\\
R256,R256,R256,R256,R256,R256,u128} \\
\texttt{u64,c7s1-3}

{\bf Discriminator architectures}
For discriminator networks, we use $70\times 70$ PatchGAN~\cite{isola2017image}. 
Let \texttt{Ck} denote a $4\times4$ Convolution-InstanceNorm-LeakyReLU layer with k filters and stride $2$. After the last layer, we apply a convolution to produce a $1$-dimensional output. We do not use InstanceNorm for the first \texttt{C64} layer. We use leaky ReLUs with a slope of $0.2$. The discriminator architecture is:\\
\texttt{C64-C128-C256-C512}

\begin{table}[ht]
    \centering
   
    \begin{tabular}{c|l}
        \noalign{\hrule height 1pt}
        layer & configuration \\
        \noalign{\hrule height 1pt}
        \multicolumn{2}{c}{Feature Generator} \\
        \noalign{\hrule height 1pt}
        1 & Conv2D (3, 64, 5, 1, 2), BN, ReLU, MaxPool \\
        \hline
        2 & Conv2D (64, 64, 5, 1, 2), BN, ReLU, MaxPool \\
        \hline
        3 & Conv2D (64, 128, 5, 1, 2), BN, ReLU \\
        \noalign{\hrule height 1pt}     
        \multicolumn{2}{c}{Disentangler} \\
        \noalign{\hrule height 1pt}
        1 & FC (8192, 3072), BN, ReLU\\
        \hline
        2 &  DropOut (0.5), FC (3072, 2048), BN, ReLU \\
        \noalign{\hrule height 1pt}     
        \multicolumn{2}{c}{Domain Identifier} \\
        \noalign{\hrule height 1pt}
        1 & FC (2048, 256), LeakyReLU \\
        \hline
        2 & FC (256, 2), LeakyReLU \\
        \noalign{\hrule height 1pt}     
        \multicolumn{2}{c}{Class Identifier} \\
        \noalign{\hrule height 1pt}
        1 & FC (2048, 10), BN, Softmax \\
        \noalign{\hrule height 1pt}     
        \multicolumn{2}{c}{Reconstructor} \\
        \noalign{\hrule height 1pt}
        1 & FC (4096, 8192) \\
        \noalign{\hrule height 1pt}     
        \multicolumn{2}{c}{Mutual Information Estimator} \\
        \noalign{\hrule height 1pt}
        fc1\_x & FC (2048, 512), LeakyReLU \\
        \hline
        fc1\_y & FC (2048, 512), LeakyReLU \\
        \hline
        2 & FC (512,1)\\
        \noalign{\hrule height 1pt}
    \end{tabular}
    \vspace{0.2cm}
    \caption{Model architecture for digit recognition task (``Digit-Five'' dataset). For each convolution layer, we list the input dimension, output dimension, kernel size, stride, and padding. For the fully-connected layer, we provide the input and output dimensions. For drop-out layers, we provide the probability of an element to be zeroed.}
     \label{tab:digit_arch}
\end{table}

\begin{table}[t]
    \centering
 
    \vspace{0.1in}
    \resizebox{\linewidth}{!}{
    \begin{tabular}{c|l}
        \noalign{\hrule height 1pt}
        layer & configuration \\
        \noalign{\hrule height 1pt}
        \multicolumn{2}{c}{Feature Generator: ResNet50 or AlexNet} \\
        \noalign{\hrule height 1pt}
        \multicolumn{2}{c}{Disentangler} \\
        \noalign{\hrule height 1pt}
        1 & Dropout(0.5), FC (2048, 2048), BN, ReLU \\
        \hline
        2 & Dropout(0.5), FC (2048, 2048), BN, ReLU \\
        \noalign{\hrule height 1pt}     
        \multicolumn{2}{c}{Domain Identifier} \\
        \noalign{\hrule height 1pt}
        1 & FC (2048, 256), LeakyReLU \\
        \hline
        2 & FC (256, 2), LeakyReLU \\
        \noalign{\hrule height 1pt}     
        \multicolumn{2}{c}{Class Identifier} \\
        \noalign{\hrule height 1pt}
        1 & FC (2048, 10), BN, Softmax \\
        \noalign{\hrule height 1pt}     
        \multicolumn{2}{c}{Reconstructor} \\
        \noalign{\hrule height 1pt}
        1 & FC (4096, 2048) \\
        \noalign{\hrule height 1pt}     
        \multicolumn{2}{c}{Mutual Information Estimator} \\
        \noalign{\hrule height 1pt}
        fc1\_x & FC (2048, 512) \\
        \hline
        fc1\_y & FC (2048, 512), LeakyReLU \\
        \hline
        2 & FC (512,1)\\
        \noalign{\hrule height 1pt}
    \end{tabular}}
    \vspace{0.2cm }
       \caption{Model Architecture and `DomainNet`. For each convolution layer, we list the input dimension, output dimension, kernel size, stride, and padding. For the fully-connected layer, we provide the input and output dimensions. For drop-out layers, we provide the probability of an element to be zeroed.}\label{tab:image_arch}
\end{table}

\end{document}